%% file: main.tex
\documentclass[10pt,twocolumn,letterpaper]{article}

\usepackage{iccv}
\usepackage{times}
\usepackage{epsfig}
\usepackage{epstopdf}
\usepackage{graphicx}
\usepackage{amsmath}
\usepackage{amssymb}
\usepackage{textcomp}
\usepackage{bbm}
\usepackage{bm}
\DeclareMathOperator*{\argmin}{arg\,min}

\usepackage[pagebackref=true,breaklinks=true,letterpaper=true,colorlinks,bookmarks=false]{hyperref}

\def\iccvPaperID{173}%

\newcommand{\JD}{}
\newcommand{\cc}{}
\newcommand{\newchanges}{}

\newcommand{\changes}{}
\newcommand{\KG}{}

\iccvfinalcopy
\ificcvfinal\fi
\begin{document}

\title{Learning image representations tied to ego-motion}

\author{Dinesh Jayaraman\\
The University of Texas at Austin\\
{\tt\small dineshj@cs.utexas.edu}
\and
Kristen Grauman\\
The University of Texas at Austin\\
{\tt\small grauman@cs.utexas.edu}
}

\maketitle

\input{abstract}
\input{intro}

\input{related}
\input{approach}
\input{exp}

\input{ack}
\newpage
\small{
\bibliographystyle{ieee}
\bibliography{refs}
}
\newpage
\input{arxiv_supp}

\end{document}

%% file: abstract.tex
\begin{abstract}
  Understanding how images of objects and scenes behave in response to specific ego-motions is a crucial aspect of proper visual development, yet existing visual learning methods are conspicuously disconnected from the physical source of their images.  We propose to exploit proprioceptive motor signals to provide unsupervised regularization in convolutional neural networks to learn visual representations from egocentric video. Specifically, we enforce that our learned features exhibit equivariance \ie they respond predictably to transformations associated with distinct ego-motions.  With three datasets, we show that our unsupervised feature learning approach significantly outperforms previous approaches on visual recognition and next-best-view prediction tasks.  In the most challenging test, we show that features learned from video captured on an autonomous driving platform improve large-scale scene recognition in static images from a disjoint domain.
\end{abstract}

%% file: intro.tex
\vspace{-0.1in}
\section{Introduction}\label{sec:intro}
\input{conceptfig}

How is visual learning shaped by ego-motion?  In their famous ``kitten carousel" experiment, psychologists Held and Hein examined this question in 1963~\cite{held1963movement}.  To analyze the role of self-produced movement in perceptual development, they designed a carousel-like apparatus in which two kittens could be harnessed.  For eight weeks after birth, the kittens were kept in a dark environment, except for one hour a day on the carousel.  One kitten, the ``active" kitten, could move freely of its own volition while attached.  The other kitten, the ``passive" kitten, was carried along in a basket and could not control his own movement; rather, he was forced to move in exactly the same way as the active kitten.  Thus, both kittens received the same visual experience.  However, while the active kitten simultaneously experienced signals about his own motor actions, the passive kitten did not.  The outcome of the experiment is remarkable.  While the active kitten's visual perception was indistinguishable from kittens raised normally, the passive kitten suffered fundamental problems.  The implication is clear: proper perceptual development requires leveraging \emph{self-generated movement in concert with visual feedback}.

We contend that today's visual recognition algorithms are crippled much like the passive kitten.  The culprit: learning from ``bags of images".  Ever since statistical learning methods emerged as the dominant paradigm in the recognition literature, the norm has been to treat images as i.i.d. draws from an underlying distribution.  Whether learning object categories, scene classes, body poses, or features themselves, the idea is to discover patterns within a collection of snapshots, blind to their physical source.  So is the answer to learn from video?  Only partially.  Without leveraging the accompanying motor signals initiated by the videographer, learning from video data does \emph{not} escape the passive kitten's predicament.

\begin{figure}[t]
  \centering
  \includegraphics[width=0.95\linewidth]{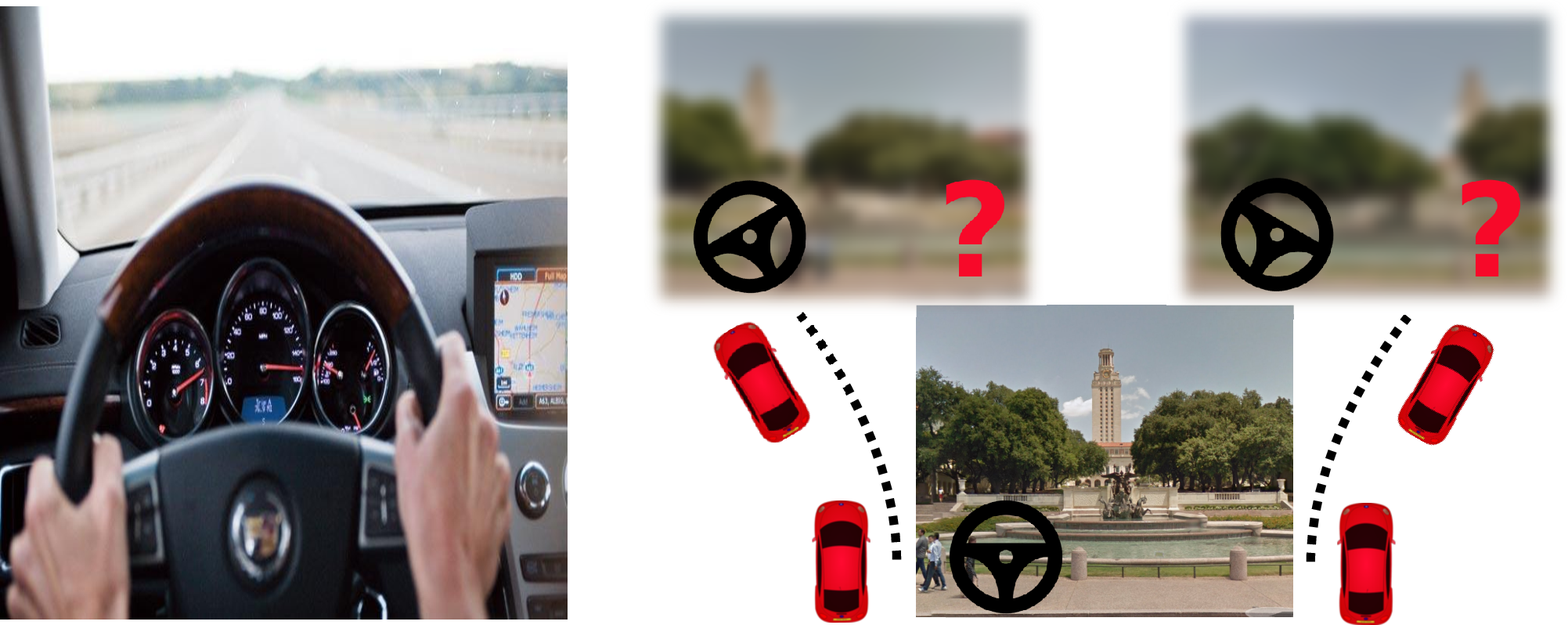}
   \caption{We learn visual features from egocentric video that respond predictably to observer egomotion.}
   \vspace{-0.1in}
  \label{fig:data_conceptfig}
\end{figure}

Inspired by this concept, we propose to treat visual learning as an embodied process, where the visual experience is inextricably linked to the motor activity behind it.\footnote{Depending on the context, the motor activity could correspond to either the 6-DOF ego-motion of the observer moving in the scene or the second-hand motion of an object being actively manipulated, e.g., by a person or robot's end effectors.} In particular, our goal is to learn representations that exploit the parallel signals of ego-motion and pixels. We hypothesize that downstream processing will benefit from a
feature space that preserves the connection between ``how I move" and ``how my visual surroundings change".

To this end, we cast the problem in terms of unsupervised equivariant feature learning.  During training, the input image sequences are accompanied by a synchronized stream of ego-motor sensor readings; \cc{however, they need not possess any semantic labels.}  The ego-motor signal could correspond, for example, to the inertial sensor measurements received alongside video on a wearable or car-mounted camera.  The objective is to learn a feature mapping from pixels in a video frame to a space that is \emph{equivariant} to various motion classes.  In other words, the learned features should \emph{change in predictable and systematic ways as a function of the transformation applied to the original input.}  See Fig~\ref{fig:conceptfig}.  We develop a convolutional neural network (CNN) approach that optimizes a feature map for the desired egomotion-based equivariance.  To exploit the features for recognition, we augment the network with a classification loss when class-labeled images are available. In this way, ego-motion serves as side information to regularize the features learned, which we show facilitates category learning when labeled examples are scarce.

In sharp contrast to our idea, previous work on visual features---whether hand-designed or learned---primarily targets feature \emph{invariance}.  Invariance is a special case of equivariance, where transformations of the input have no effect.  Typically, one seeks invariance to small transformations, e.g., the orientation binning and pooling operations in SIFT/HOG and modern CNNs both target invariance to local translations and rotations.  While a powerful concept, invariant representations require a delicate balance: ``too much" invariance leads to a loss of useful information or discriminability.  In contrast, more general equivariant representations are intriguing for their capacity to impose structure on the output space without forcing a loss of information. Equivariance is ``active'' in that it exploits observer motor signals, like Hein and Held's active kitten. %

Our main contribution is a novel feature learning approach that couples ego-motor signals and video.  To our knowledge, ours is the first attempt to ground feature learning in physical activity.
The limited prior work on unsupervised feature learning with video~\cite{Mobahi2009,Ranzato2014, Michalski2014,Goroshin2014} learns only passively from observed scene dynamics, uninformed by explicit motor sensory cues.  Furthermore, while equivariance is explored in some recent work, unlike our idea, it typically focuses on 2D image transformations as opposed to 3D ego-motion~\cite{kivinen2011transformation,schmidt2012learning} and considers existing features~\cite{tinne-survey,Vedaldi2014}.  Finally, whereas existing methods that learn from image transformations focus on view synthesis applications~\cite{Hinton2011,kulkarni2015deep,Michalski2014}, we explore recognition applications of learning jointly equivariant and discriminative feature maps.

We apply our approach to three public datasets.  On pure equivariance as well as recognition tasks, our method consistently outperforms the most related techniques in feature learning.  In the most challenging test of our method, we show that features learned from video captured on a vehicle can improve image recognition accuracy on a disjoint domain.  In particular, we use unlabeled KITTI~\cite{kitti, kitti1} car data to regularize feature learning for the 397-class scene recognition task for the SUN dataset~\cite{sun}.  Our results show the promise of departing from the ``bag of images" mindset, in favor of an embodied approach to feature learning.

%% file: conceptfig.tex
\begin{figure*}[t]
  \centering
  \includegraphics[width=1\linewidth]{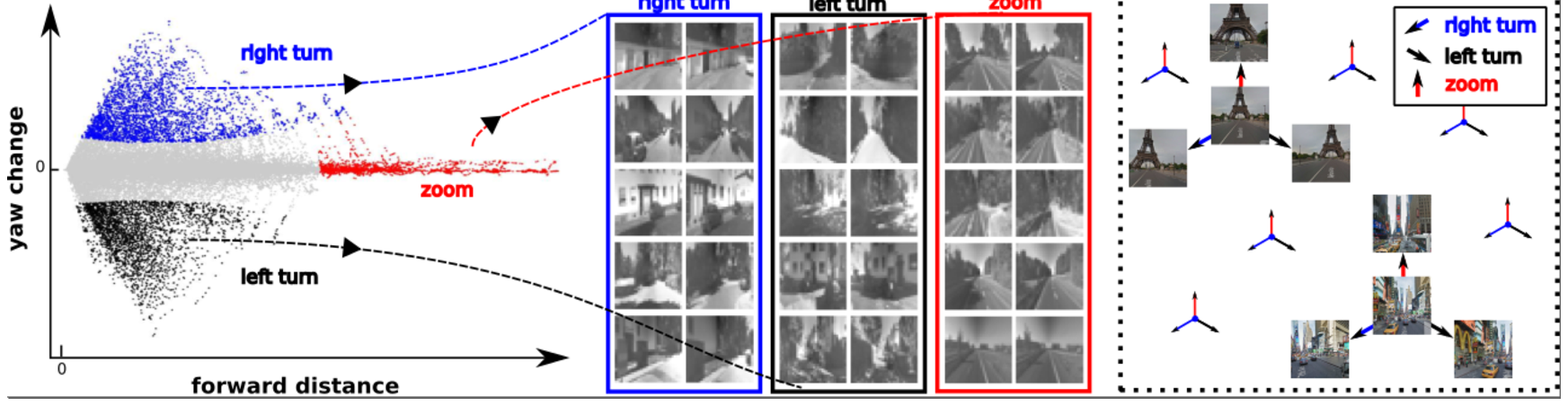}
  \caption{Our goal is to learn a feature space equivariant to ego-motion.  We train with image pairs from video accompanied by their sensed ego-poses (left and center), and produce a feature mapping such that two images undergoing the same ego-pose \emph{change} move similarly in the feature space (right).  \textbf{Left:}  Scatter plot of motions $(\bm{y}_i - \bm{y}_j)$ among pairs of frames $\leq$ 1s apart in video from KITTI car-mounted camera, clustered into motion patterns $p_{ij}$. \textbf{Center:} Frame pairs $(\bm{x}_i,\bm{x}_j)$ from the ``right turn'', ``left turn'' and ``zoom'' motion patterns. \textbf{Right:} An illustration of the equivariance property we seek in the learned feature space. Pairs of frames corresponding to each ego-motion pattern ought to have predictable relative positions in the learned feature space.  Best seen in color.}
\vspace{-0.1in}
  \label{fig:conceptfig}
\end{figure*}

%% file: related.tex
\vspace{-0.05in}
\section{Related work}\label{sec:related}

\paragraph{Invariant features}

Invariance is a special case of equivariance, wherein a transformed output remains identical to its input.  Invariance is known to be valuable for visual representations.  Descriptors like SIFT, HOG, and aspects of CNNs like pooling and convolution, are hand-designed for invariance to small shifts and rotations.  Feature learning work aims to \emph{learn} invariances from data~\cite{Simard1998,simard2003best,vincent2008extracting,sohn2012learning,Dosovitskiy2014}.  Strategies include augmenting training data by perturbing image instances with label-preserving transformations~\cite{simard2003best,vincent2008extracting,Dosovitskiy2014}, and inserting linear transformation operators into the feature learning algorithm~\cite{sohn2012learning}.

Most relevant to our work are feature learning methods based on temporal coherence and ``slow feature analysis"~\cite{Wiskott2002,drlim,Mobahi2009}.  The idea is to require that learned features vary slowly over continuous video, since
visual stimuli can only gradually change between adjacent frames.  Temporal coherence has been explored for unsupervised feature learning with CNNs~\cite{Mobahi2009,zou2012deep,Goroshin2014,cadieu2012learning,lies2014slowness}, with applications to dimensionality reduction~\cite{drlim}, object  recognition~\cite{Mobahi2009,zou2012deep}, and metric learning~\cite{Goroshin2014}. Temporal coherence of inferred body poses in unlabeled video is exploited for invariant recognition in~\cite{chaoyeh}.
These methods exploit video as a source of free supervision to achieve invariance, analogous to the image perturbations idea above.  In contrast, our method exploits video coupled with ego-motor signals to achieve the more general property of equivariance.  %

\vspace{-0.1in}
\paragraph{Equivariant representations}
Equivariant features can also be hand-designed or learned.  For example, equivariant or ``co-variant" operators are designed to detect repeatable interest points~\cite{tinne-survey}.  Recent work explores ways to learn descriptors with in-plane translation/rotation equivariance~\cite{kivinen2011transformation,schmidt2012learning}.  While the latter does perform feature learning, its equivariance properties are crafted for specific 2D image transformations.  In contrast, we target more complex equivariances arising from natural observer motions (3D ego-motion) that cannot easily be crafted, and our method learns them from data.

\newchanges{
{Methods to learn representations} with disentangled latent factors~\cite{Hinton2011,kulkarni2015deep} aim to sort properties like pose, illumination \etc into distinct portions of the feature space.  For example, the transforming auto-encoder learns to explicitly represent instantiation parameters of object parts in equivariant hidden layer units~\cite{Hinton2011}.
Such methods target equivariance in the limited sense of inferring pose parameters, which are appended to a conventional feature space designed to be invariant.  In contrast, our formulation encourages \emph{equivariance} over the \emph{complete} feature space; we show the impact as an unsupervised regularizer when training a recognition model with limited training data.
}

{The work of~\cite{Vedaldi2014} quantifies the invariance/equivariance of
various standard representations, including CNN features, in terms of their responses to specified in-plane 2D image transformations (affine warps, flips of the image).}   We adopt the definition of equivariance used in that work, but our goal is entirely different.  Whereas~\cite{Vedaldi2014} quantifies the equivariance of existing descriptors, our approach learns a feature space that is equivariant.

\vspace{-0.1in}
\paragraph{Learning transformations}

Other methods train with pairs of transformed images and infer an implicit representation for the transformation itself. In~\cite{Memisevic2013},
bilinear models with multiplicative interactions are used to learn content-independent ``motion features" that encode only the transformation between image pairs.
One such model, the ``gated autoencoder'' is extended to perform sequence prediction for video in~\cite{Michalski2014}. %
Recurrent neural networks combined  with a grammar model of scene dynamics can also predict future frames in video~\cite{Ranzato2014}.  Whereas these methods learn a representation for image pairs (or tuples) related by some transformation, we learn a representation for individual images in which the behavior under transformations is predictable.  Furthermore, whereas these prior methods abstract away the image content, our method preserves it, making our features relevant for recognition.

\vspace{-0.05in}
\paragraph{Egocentric vision}

There is renewed interest in egocentric computer vision methods, though none perform feature learning using motor signals and pixels in concert as we propose.  Recent methods use ego-motion cues to separate foreground and background~\cite{ren-cvpr2010,xu2012moving} or infer the first-person gaze~\cite{yamada2012attention,li2013iccv}.  While most work relies solely on apparent image motion, the method of~\cite{xu2012moving} exploits a robot's motor signals to detect moving objects and~\cite{nakamura1995motion} uses reinforcement learning to form robot movement policies by exploiting correlations between motor commands and observed motion cues.%

%% file: approach.tex
\section{Approach}\label{sec:approach}

Our goal is to learn an image representation that is equivariant with respect to ego-motion transformations.
Let $\bm{x}_i \in \mathcal{X}$ be an image in the original pixel space, and let $\bm{y}_i \in \mathcal{Y}$ be its associated ego-pose representation.  The ego-pose captures the available motor signals, and could take a variety of forms.  For example, $\mathcal{Y}$ may encode the complete observer camera pose (its position in 3D space, pitch, yaw, roll), some subset of those parameters, or any reading from a motor sensor paired with the camera.

As input to our learning algorithm, we have a training set $\mathcal{U}$ of $N_u$ image pairs and their associated ego-poses, $\mathcal{U} = \{\langle (\bm{x}_i,\bm{x}_j), (\bm{y}_i, \bm{y}_j) \rangle\}_{(i,j)=1}^{N_u}$.
The image pairs originate from video sequences, though they need not be adjacent frames in time.   The set may contain pairs from multiple videos and cameras.  Note that this training data does \emph{not} have any semantic labels (object categories, \etc); they are ``labeled" only in terms of the ego-motor sensor readings.

In the following, we first explain how to translate ego-pose information into pairwise ``motion pattern'' annotations (Sec~\ref{sec:motionPatterns}).  Then, Sec~\ref{sec:equivar} defines the precise nature of the equivariance we seek, and Sec~\ref{sec:loss_func} defines our learning objective. Sec~\ref{sec:recog} shows how our equivariant feature learning scheme may be used to enhance recognition with limited training data. Finally, in Sec~\ref{sec:siamese}, we show how a feedforward neural network architecture may be trained to produce the desired equivariant feature space.

\subsection{Mining discrete ego-motion patterns} \label{sec:motionPatterns}

First we want to organize training sample pairs into a discrete set of ego-motion patterns.  For instance, one ego-motion pattern might correspond to ``tilt downwards by approximately 20\textdegree''.
While one could collect new data explicitly controlling for the patterns (e.g., with a turntable and camera rig), we prefer a data-driven approach that can leverage video and ego-pose data collected ``in the wild''.

To this end, we discover clusters among pose difference vectors $\bm{y}_i-\bm{y}_j$ for pairs $(i,j)$ of temporally close frames from video (typically $\lessapprox$1 second apart; see Sec~\ref{sec:setup} for details).  For simplicity we apply $k$-means to find $G$ clusters, though other methods are possible.  Let $p_{ij}\in\mathcal{P}=\{1,\dots,G\}$ denote the motion pattern ID, \ie, the cluster to which $(\bm{y}_i,\bm{y}_j)$ belongs.  We can now replace the ego-pose vectors in $\mathcal{U}$ with motion pattern IDs: $\langle (\bm{x}_i,\bm{x}_j), p_{ij} \rangle$.
\footnote{For movement with $d$ degrees of freedom, setting $G\approx d$ should suffice (cf. Sec~\ref{sec:equivar}). We chose small $G$ for speed and did not vary it.}

The left panel of Fig~\ref{fig:conceptfig} illustrates a set of motion patterns discovered from videos in the KITTI~\cite{kitti} dataset, which are captured from a moving car.  Here $\mathcal{Y}$ consists of the position and yaw angle of the camera.  So, we are clustering a 2D space consisting of forward distance and change in yaw.  As illustrated in the center panel, the largest clusters correspond to the car's three primary ego-motions: turning left, turning right, and going forward.

\subsection{Ego-motion equivariance}\label{sec:equivar}

Given $\mathcal{U}$, we wish to learn a feature mapping function $\mathbf{z}_{\bm{\theta}}(.): \mathcal{X} \rightarrow \mathcal{R}^D$ parameterized by $\bm{\theta}$ that maps a single image to a $D$-dimensional vector space that is equivariant to ego-motion.  To be equivariant, the function $\mathbf{z}_{\bm{\theta}}$ must respond \emph{systematically} and \emph{predictably} to ego-motion:
\vspace{-0.05in}
\begin{equation}
  \mathbf{z}_{\bm{\theta}}(\bm{x}_j) \approx f(\mathbf{z}_{\bm{\theta}}(\bm{x}_i),\bm{y}_i,\bm{y}_j),
\end{equation}
for some function $f$.  We consider equivariance for linear \changes{functions $f(.)$}, following~\cite{Vedaldi2014}.  In this case,
$\mathbf{z}_{\bm{\theta}}$ is said to be equivariant with respect to some transformation $g$  if there exists a $D\times D$ matrix\footnote{bias dimension assumed to be included in $D$ for notational simplicity} $M_g$ such that:
\vspace{-0.05in}
\begin{equation}
  \forall \bm{x} \in \mathcal{X}:\mathbf{z}_{\bm{\theta}}(g \bm{x}) \approx M_g \mathbf{z}_{\bm{\theta}}(\bm{x}).
  \label{eq:equivar}
\end{equation}
Such an $M_g$ is called the ``equivariance map'' of $g$ on the feature space $\mathbf{z}_{\bm{\theta}}(.)$.  It represents the affine transformation in the feature space that corresponds to transformation $g$ in the pixel space.  For example, suppose a motion pattern $g$ corresponds to a yaw turn of 20\textdegree, and $\bm{x}$ and  $g\bm{x}$ are the images observed before and after the turn, respectively.  Equivariance demands that there is some matrix $M_g$ that maps the pre-turn image to the post-turn image, once those images are expressed in the feature space $\mathbf{z}_{\bm{\theta}}$.  Hence, $\mathbf{z}_{\bm{\theta}}$ ``organizes" the feature space in such a way that movement in a particular direction in the feature space (here, as computed by multiplication with $M_g$) has a predictable outcome.  The linear case, as also studied in~\cite{Vedaldi2014}, ensures that the structure of the mapping has a simple form, and is convenient for learning since $M_g$ can be encoded as a fully connected layer in a neural network.

While prior work{~\cite{kivinen2011transformation,schmidt2012learning}} focuses on equivariance where $g$ is a 2D image warp, we explore {the case}
{where $g\in\mathcal{P}$ is an ego-motion pattern (cf. Sec~\ref{sec:motionPatterns}) reflecting the observer's 3D movement in the world.}
In theory, appearance changes of an image in response to an observer's ego-motion are not determined by the ego-motion alone.  They also depend on the depth map of the scene and the motion of dynamic objects in the scene.  One could easily augment either the frames $\bm{x}_i$ or the ego-pose $\bm{y}_i$ with depth maps, when available.  Non-observer motion appears more difficult, especially in the face of changing occlusions and newly appearing objects.  However, our experiments indicate we can learn effective representations even with dynamic objects.  In our implementation, we train with pairs relatively close in time, so as to avoid some of these pitfalls.

While during training {we target equivariance for the discrete set of $G$ ego-motions}, the learned feature space will \emph{not} be limited to preserving equivariance for pairs originating from the same ego-motions.  This is because the linear equivariance maps are composable.  If we are operating in a space where every ego-motion can be composed as a sequence of ``atomic'' motions, equivariance to those atomic motions is sufficient to guarantee equivariance to all motions.   To see this, suppose that the maps for ``turn head right by 10\textdegree'' (ego-motion pattern $r$) and ``turn head up by 10\textdegree'' (ego-motion pattern $u$) are respectively $M_r$ and $M_u$, \ie, $\mathbf{z}(r \bm{x})=M_r \mathbf{z}(\bm{x})$ and $\mathbf{z}(u \bm{x})=M_u \mathbf{z}(\bm{x})$ for all $\bm{x}\in \mathcal{X}$.  Now for a novel diagonal motion $d$ that can be composed from these atomic motions as $d=r \circ u$, we have
\vspace{-0.05in}
\begin{align}
  \mathbf{z}(d\bm{x})&=\mathbf{z}((r \circ u) \bm{x})=M_r \mathbf{z}(u\bm{x})=M_r M_u \mathbf{z}(\bm{x}),
  \label{eq:motion_bases}
\end{align}
{so that $M_d=M_rM_u$} is the equivariance map for novel ego-motion $d$, even though $d$ was not among $1,\dots,G$.  This property lets us {restrict our attention to
a relatively small number of discrete ego-motion patterns} during training, and still learn features equivariant w.r.t. new ego-motions.

\subsection{Equivariant feature learning objective}\label{sec:loss_func}

We now design a loss function that encourages the learned feature space $\mathbf{z}_{\bm{\theta}}$ to exhibit equivariance with respect to each ego-motion pattern. Specifically, we would like to learn the optimal feature space parameters $\bm{\theta^*}$ jointly with its equivariance maps $\mathcal{M}^*=\{M_1^*,\dots,M_G^*\}$ for the motion pattern clusters $1$ through $G$ (cf. Sec~\ref{sec:motionPatterns}).

To achieve this, a naive translation of the definition of equivariance in Eq~\eqref{eq:equivar} into a minimization problem over feature space parameters $\bm{\theta}$ and the $D\times D$ equivariance map candidate matrices $\mathcal{M}$ would be as follows:
\vspace{-0.05in}

\begin{equation}
  (\bm{\theta}^*, \mathcal{M}^*) = \argmin_{\bm{\theta}, \mathcal{M}} \sum_g \sum_{ \changes{\{(i,j):p_{ij}=g\}}} d\left(\changes{M_g \mathbf{z}_{\bm{\theta}}(\bm{x}_i), \mathbf{z}_{\bm{\theta}}(\bm{x}_j)}\right),
  \label{eq:loss0}
\end{equation}
where $d(.,.)$ is a distance measure. This problem can be decomposed into $G$ independent optimization problems, one for each motion, corresponding only to the inner summation above, and dealing with disjoint data. The $g$-th such problem requires only that training frame pairs annotated with motion pattern $p_{ij} = g$ approximately satisfy Eq~\eqref{eq:equivar}.

However, such a formulation admits problematic solutions that perfectly optimize it, \eg for the trivial all-zero feature space $\mathbf{z}_{\bm{\theta}}(\bm{x})=\bm{0}, \forall \bm{x} \in \mathcal{X}$ with $M_g$ set to the all-zeros matrix for all $g$, the loss above evaluates to zero. To avoid such solutions, and to force the learned $M_g$'s to be different from one another (since we would like the learned representation to respond \emph{differently} to different ego-motions), we simultaneously account for the ``negatives" of each motion pattern.  Our learning objective is:
\begin{equation}
  (\bm{\theta}^*, \mathcal{M}^*) = \argmin_{\bm{\theta}, \mathcal{M}} \sum_{g,i,j} d_g\left(\changes{M_g \mathbf{z}_{\bm{\theta}}(\bm{x}_i),\mathbf{z}_{\bm{\theta}}(\bm{x}_j)}, p_{ij}\right),
  \label{eq:loss1}
\end{equation}
where $d_g(.,.,.)$ is a ``contrastive loss''~\cite{drlim} specific to motion pattern $g$:
\begin{multline}
  d_g(\bm{a},\bm{b},c)=\mathbbm{1}(c=g) d(\bm{a},\bm{b})+\\
  \mathbbm{1}(c\neq g) \max(\delta-d(\bm{a},\bm{b}),0),
  \label{eq:contrastive_loss}
\end{multline}
where $\mathbbm{1}(.)$ is the indicator function.  This contrastive loss penalizes distance between $\bm{a}$ and $\bm{b}$ in ``positive'' mode (when $c=g$), and pushes apart pairs in ``negative'' mode (when $c \neq g$), up to a minimum margin distance specified by the constant $\delta$. We use the $\ell_2$ norm for the distance $d(.,.)$.

In our objective in Eq~\eqref{eq:loss1}, the contrastive loss operates in the latent feature space. For pairs belonging to cluster $g$, the contrastive loss $d_g$ penalizes feature space distance between the first image and its transformed pair, similar to Eq~\eqref{eq:loss0} above.  For pairs belonging to clusters other than $g$, $d_g$ requires that the transformation defined by $M_g$ must not bring the image representations close together.  In this way, our objective learns the $M_g$'s jointly. It ensures that distinct ego-motions, when applied to an input $\mathbf{z}_{\bm{\theta}}(\bm{x})$, map it to different locations in feature space.

We want to highlight the important distinctions between our objective and the ``temporal coherence'' objective of \cite{Mobahi2009} \cc{for slow feature analysis}.   Written in our notation, the objective of \cite{Mobahi2009} may be stated as:
\vspace{-0.05in}
\begin{equation}
  \bm{\theta}^* = \argmin_{\bm{\theta}} \sum_{i,j} d_{1}(\mathbf{z}_{\bm{\theta}}(\bm{x}_i),\mathbf{z}_{\bm{\theta}}(\bm{x}_j), \mathbbm{1}(|t_i-t_j|\leq T)),
  \label{eq:temp_coherence}
\end{equation}
where $t_i, t_j$ are the video time indices of $\bm{x}_i$, $\bm{x}_j$ and $T$ is a temporal neighborhood size hyperparameter.  This loss encourages the representations of nearby frames to be similar to one another.  However, crucially, it does not account for the nature of the ego-motion between the frames.  Accordingly, while temporal coherence helps learn invariance to small image changes, it does not target a (more general) equivariant space.  Like the passive kitten from Hein and Held's experiment, the temporal coherence constraint watches video to passively learn a representation; like the active kitten, our method registers the \emph{observer motion} explicitly with the video to learn more effectively, as we will demonstrate in results.

\subsection{Regularizing a recognition task}\label{sec:recog}

While we have thus far described our formulation for generic equivariant image representation learning, it can optionally be used for visual recognition tasks.
Suppose that in addition to the ego-pose annotated pairs $\mathcal{U}$ we are also given a small set of $N_l$ class-labeled \newchanges{static} images, $\mathcal{L} = \{(\bm{x}_k,c_k\}_{k=1}^{N_l}$, where $c_k \in \{1,\dots,C\}$.
Let $L_e$ denote the unsupervised equivariance loss of Eq~\eqref{eq:loss1}. We can integrate our unsupervised feature learning scheme with the recognition task, by optimizing a misclassification loss together with $L_e$.
Let $W$ be a $D \times C$ matrix of classifier weights. {We solve jointly for $W$ and the maps $\mathcal{M}$}:
\begin{equation}
  (\bm{\theta}^*,W^*,\mathcal{M}^*)=\argmin_{\bm{\theta},W,\mathcal{M}} L_{c}(\changes{\bm{\theta}}, W,\mathcal{L})  + \lambda L_{e}(\bm{\theta}, \mathcal{M},\mathcal{U}),
  \label{eq:recog}
\end{equation}
where $L_c$ denotes the softmax loss over the learned features, $L_c(W,\mathcal{L}) = -\frac{1}{N_l} \sum_{i=1}^{N_l} \log(\sigma_{c_k}(W\mathbf{z}_{\bm{\theta}}(\bm{x}_i))$, and $\sigma_{c_k}(.)$ is the softmax probability of the correct class.
The regularizer weight $\lambda$ is a hyperparameter. \newchanges{Note that neither the supervised training data $\mathcal{L}$ nor the testing data for recognition are required to have any associated sensor data. Thus, our features are applicable to standard image recognition tasks.}

In this use case, the unsupervised ego-motion equivariance loss encodes a prior over the feature space that can improve performance on the supervised recognition task with limited training examples. \changes{We hypothesize that a feature space that embeds knowledge of how objects change under different viewpoints / manipulations allows a recognition system to, in some sense, hallucinate new views of an object to improve performance.}

\subsection{Form of the feature mapping function $\mathbf{z}_{\bm{\theta}}(.)$}\label{sec:siamese}

\begin{figure}[t]
  \centering
  \includegraphics[width=0.95\linewidth]{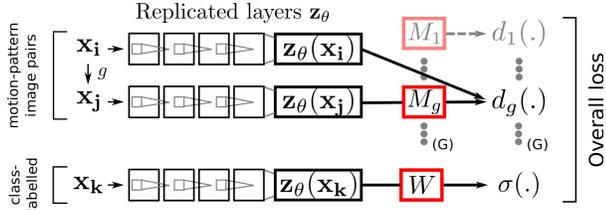}
  \caption{Training setup: (top) ``Siamese network'' for computing the equivariance loss of Eq~\eqref{eq:loss1}, together with (bottom) a third tied stack for computing the supervised recognition softmax loss as in Eq~\eqref{eq:recog}.  See Sec~\ref{sec:setup} and Supp for exact network specifications.
}
\vspace{-0.05in}
  \label{fig:siamese}
\end{figure}

For the mapping $\mathbf{z}_{\bm{\theta}}(.)$, we use a convolutional neural network architecture, so that the parameter vector $\bm{\theta}$ now represents the layer weights. The loss $L_e$ of Eq~\eqref{eq:loss1} is optimized by sharing the weight parameters $\bm{\theta}$ among two identical stacks of layers in a ``Siamese'' network~\cite{Bromley1993,drlim,Mobahi2009}, as shown in the top two rows of Fig~\ref{fig:siamese}. {Image pairs from $\mathcal{U}$ are fed into these two stacks.} Both stacks are initialized with identical random weights, and identical gradients are passed through them in every training epoch, so that the weights remain tied throughout.  Each stack encodes the feature map that we wish to train, $\mathbf{z}_{\bm{\theta}}$.

To optimize Eq~\eqref{eq:loss1}, an array of equivarance maps $\mathcal{M}$, each represented by a fully connected layer, is connected to the top of the second stack. Each such equivariance map then feeds into a motion-pattern-specific contrastive loss function $d_{g}$, whose other inputs are the first stack output and the ego-motion pattern ID $p_{ij}$.

To optimize Eq~\eqref{eq:recog}, in addition to the Siamese net that minimizes $L_{e}$ as above, the supervised softmax loss is minimized through a third replica of the $\mathbf{z}_{\bm{\theta}}$ layer stack with weights tied to the two Siamese networks stacks. {Labelled images from $\mathcal{L}$ are fed into this stack, and its output is fed into} a softmax layer whose other input is the class label. The complete scheme is depicted in Fig~\ref{fig:siamese}.
Optimization is done through mini-batch stochastic gradient descent implemented through backpropagation with the Caffe package~\cite{caffe} (more details in Sec~\ref{sec:exp} and Supp).

%% file: exp.tex
\section{Experiments}\label{sec:exp}
\begin{table*}[ht]
  \centering
    \small{
    \begin{tabular}{|l|cc|cccc|c|}
    \hline
    Tasks$\rightarrow$ & \multicolumn{2}{c|}{Equivariance error} &  \multicolumn{4}{c|}{Recognition accuracy \%} & {Next-best view} \\
    Datasets$\rightarrow$         &  \multicolumn{2}{c|}{NORB}  &  NORB-NORB     &  KITTI-KITTI   & KITTI-SUN & KITTI-SUN  & {NORB} \\
    Methods$\downarrow$          & atomic       &  composite &   [25 cls]    &  [4 cls]       & [397 cls] & [397 cls, top-10]& 1-view$\rightarrow$ 2-view \\ \hline
    random           & 1.0000       & 1.0000     &   4.00        &   25.00        &   0.25    &   2.52    & 4.00 $\rightarrow$ 4.00 \\
    \textsc{clsnet}&0.9239 & 0.9145                         &25.11$\pm$0.72 & 41.81$\pm$0.38 & 0.70$\pm$0.12  & 6.10$\pm$0.67  & - \\
    \textsc{temporal}~\cite{Mobahi2009}&0.7587 & 0.8119     &35.47$\pm$0.51 & 45.12$\pm$1.21 & 1.21$\pm$0.14  & 8.24$\pm$0.25  & 29.60$\rightarrow$ 31.90 \\
    \textsc{drlim}~\cite{drlim}&0.6404 & 0.7263             &36.60$\pm$0.41 & 47.04$\pm$0.50 & 1.02$\pm$0.12  & 6.78$\pm$0.32  & 14.89$\rightarrow$  17.95 \\
    \textsc{equiv}& \textbf{0.6082} & \textbf{0.6982}                          &\textbf{38.48$\pm$0.89} & \textbf{50.64$\pm$0.88} & \textbf{1.31$\pm$0.07} & \textbf{8.59$\pm$0.16} & \textbf{38.52}\textbf{$\rightarrow$}\textbf{43.86}\\
    \textsc{equiv+drlim}& \textbf{0.5814} & \textbf{0.6492} &\textbf{40.78$\pm$0.60} &\textbf{50.84$\pm$0.43}  &\textbf{1.58$\pm$0.17} & \textbf{9.57$\pm$0.32} & \textbf{38.46}\textbf{$\rightarrow$}\textbf{43.18} \\ \hline
    \end{tabular}
}
\caption{(Left) Average equivariance error (Eq~\eqref{eq:equiv_measure}) on NORB for ego-motions like those in the training set (atomic) and novel ego-motions (composite). (Center) Recognition result for 3 datasets (mean $\pm$ standard error) of accuracy \% over 5 repetitions. (Right) Next-best view selection accuracy \%. Our method \textsc{equiv} (and augmented with slowness in \textsc{equiv+drlim}) clearly outperforms all baselines.}
\vspace{-0.05in}
\label{tab:equiv_measurement}
\label{tab:recog}
\label{tab:nbv}
\end{table*}

We validate our approach on 3 public datasets and compare to two existing methods, on equivariance (Sec~\ref{sec:sanitycheck}), recognition performance (Sec~\ref{sec:recog_results}) and next-best view selection (Sec~\ref{sec:nbv}).
Throughout we compare the following methods:
\begin{itemize}
\vspace{-0.05in}
\item \textsc{clsnet}: A neural network trained only from the supervised samples with a softmax loss.
\vspace{-0.05in}
\item \textsc{temporal}: The \emph{temporal coherence} approach of~\cite{Mobahi2009}, which regularizes the classification loss with Eq~\eqref{eq:temp_coherence} \JD{setting the distance measure $d(.)$ to the $\ell_1$ distance in $d_{1}$}.  This method aims to learn invariant features by exploiting the fact that adjacent video frames should not change too much.
\vspace{-0.05in}
\item \textsc{drlim}: The approach of~\cite{drlim}, which also regularizes the classification loss with Eq~\eqref{eq:temp_coherence}, but \JD{setting $d(.)$ to the $\ell_2$ distance in $d_{1}$}.
\vspace{-0.05in}
\item \textsc{equiv}: Our ego-motion equivariant feature learning approach, combined with the classification loss as in Eq~\eqref{eq:recog}, unless otherwise noted below.
\vspace{-0.05in}
\item \textsc{equiv+drlim}: Our approach augmented with temporal coherence regularization (\cite{drlim}).
\vspace{-0.05in}
\end{itemize}

{\textsc{temporal} and \textsc{drlim} are the most pertinent baselines because they, like us, use contrastive loss-based formulations, but represent the popular ``slowness''-based family of techniques (\cite{zou2012deep, cadieu2012learning, Goroshin2014, lies2014slowness}) for unsupervised feature learning from video, which, unlike our approach, are passive.}

\subsection{Experimental setup details} \label{sec:setup}

Recall that in the fully unsupervised mode, our method trains with pairs of video frames annotated only by their ego-poses in $\mathcal{U}$.  In the supervised mode, when applied to recognition, our method additionally has access to a set of class-labeled images in $\mathcal{L}$.  Similarly, the baselines all receive a pool of unsupervised data and supervised data.  We now detail the data composing these two sets.
\vspace*{-0.1in}
\paragraph{Unsupervised datasets} We consider two unsupervised datasets, NORB and KITTI:

\noindent
\newchanges{
(1) \textbf{NORB}~\cite{norb}: This dataset has 24,300 96$\times$96-pixel images of 25 toys captured by systematically varying camera pose. We generate a random 67\%-33\% train-validation split and use 2D ego-pose vectors $\bm{y}$ consisting of camera elevation and azimuth.  Because this dataset has discrete ego-pose variations, we consider two ego-motion patterns, \ie, $G=2$ (cf.~Sec~\ref{sec:motionPatterns}): one step along elevation and one step along azimuth. For \textsc{equiv}, we use all available positive pairs for each of the two motion patterns from the training images, yielding a $N_u=45,417$-pair training set. For \textsc{drlim} and \textsc{temporal}, we create a 50,000-pair training set (positives to negatives ratio 1:3).  Pairs within one step (elevation and/or azimuth) are treated as ``temporal neighbors", as in the turntable results of~\cite{drlim,Mobahi2009}.
}

\noindent
\newchanges{
(2) \textbf{KITTI}~\cite{kitti,kitti1}: This dataset contains videos with registered GPS/IMU sensor streams captured on a car driving around 4 types of areas (location classes): ``campus'', ``city'', ``residential'', ``road''. We generate a random 67\%-33\% train-validation split and use 2D ego-pose vectors consisting of ``yaw'' and ``forward position'' (integral over ``forward velocity'' sensor outputs) from the sensors.  We discover ego-motion patterns $p_{ij}$ (cf.~Sec~\ref{sec:motionPatterns}) on frame pairs $\leq$ 1 second apart. We compute $6$ clusters and automatically retain the $G=3$ with the largest motions, which upon inspection correspond to ``forward motion/zoom", ``right turn", and ``left turn" (see Fig~\ref{fig:conceptfig}, left). For \textsc{equiv}, we create a $N_u=47,984$-pair training set {with 11,996 positives}.  %
For \textsc{drlim} and \textsc{temporal}, we create a 98,460-pair training set with 24,615 ``temporal neighbor'' positives sampled $\leq$2 seconds apart. We use grayscale ``camera 0'' frames (see~\cite{kitti1}), downsampled to 32$\times$32 pixels, \KG{so that we can adopt CNN architecture choices known to be effective for tiny images~\cite{cuda-convnet}.}
}

\vspace*{-0.1in}
\paragraph{Supervised datasets}

In our recognition experiments, we consider 3 supervised datasets $\mathcal{L}$: (1) \textbf{NORB}: We select 6 images from each of the $C=25$ object training splits at random to create instance recognition training data.  (2) \textbf{KITTI}: We select 4 images from each of the $C=4$ location class training splits at random to create location recognition training data.(3) \textbf{SUN}~\cite{sun}: We select 6 images for each of $C=397$ scene categories at random to create scene recognition training data.  We preprocess them identically to the KITTI images above (grayscale, crop to KITTI aspect ratio, resize to $32 \times 32$).  We keep all the supervised datasets small, since unsupervised feature learning should be most beneficial when labeled data is scarce. \newchanges{Note that while the video frames of the unsupervised datasets $U$ are associated with ego-poses, the static images of $\mathcal{L}$ have no such auxiliary data.}

\vspace*{-0.1in}
\paragraph{Network architectures and optimization}

\newchanges{
For KITTI, we closely follow the cuda-convnet~\cite{cuda-convnet} recommended CIFAR-10 architecture: 32 conv(5x5)-max(3x3)-ReLU $\rightarrow$ 32 conv(5x5)-ReLU-avg(3x3) $\rightarrow$ 64 conv(5x5)-ReLU-avg(3x3) $\rightarrow$ $D=$64 full feature units. For NORB, we use a fully connected architecture: 20 full-ReLU$\rightarrow$ $D=$100 full feature units. \cc{Parentheses indicate sizes of convolution or pooling kernels, and pooling layers have stride length 2.}
}

  We use Nesterov-accelerated stochastic gradient descent.
 The base learning rate and regularization $\lambda$s are selected with greedy cross-validation.
The contrastive loss margin parameter $\delta$ in Eq~\eqref{eq:contrastive_loss} is set to 1.0.
We report all results for all methods based on 5 repetitions.  For more details on architectures and optimization, see Supp.

\subsection{Equivariance measurement}\label{sec:sanitycheck}

First, we test the learned features for equivariance.  %
Equivariance is measured separately for each ego-motion $g$ through the normalized error $\rho_g$:\vspace{-0.05in}
\begin{equation}
  \rho_g=E\left[{\|\mathbf{z_{\bm{\theta}}}(\bm{x})-M_g^{'} \mathbf{z_{\bm{\theta}}}(g\bm{x})\|_2}/{\|\mathbf{z_{\bm{\theta}}}(\bm{x})-\mathbf{z_{\bm{\theta}}}(g\bm{x})\|_2}\right],
  \label{eq:equiv_measure}
\end{equation}
where $E[.]$ denotes the empirical mean, $M_g^{'}$ is the equivariance map, and $\rho_g=0$ would signify perfect equivariance. We closely follow the equivariance evaluation approach of~\cite{Vedaldi2014} to solve for the equivariance maps of features produced by each compared method on held-out validation data, \cc{before computing $\rho_g$} (see Supp).

{
We test both (1) ``atomic'' ego-motions matching those provided in the training pairs (\ie, ``up" 5\textdegree and ``down" 20\textdegree) and (2) composite ego-motions (``up+right'', ``up+left'', ``down+right'').  \cc{The latter lets us verify that our method's equivariance extends beyond those motion patterns used for training (cf.~Sec~\ref{sec:equivar}).}
}
First, as a sanity check, we quantify equivariance for the unsupervised loss of Eq~\eqref{eq:loss1} in isolation, \ie, learning with only $\mathcal{U}$.  Our \textsc{equiv} method's average $\rho_g$ error is 0.0304 and 0.0394 for atomic and composite ego-motions in NORB, respectively.  In comparison, \textsc{drlim}---which promotes invariance, not equivariance---achieves $\rho_g =$ 0.3751 and 0.4532.  Thus, \cc{without class supervision,} \textsc{equiv} tends to learn nearly completely equivariant features, even for novel composite transformations.

Next we evaluate equivariance for all methods using features optimized for the NORB recognition task.
Table~\ref{tab:equiv_measurement} (left) shows the results. \newchanges{As expected, we find that the features learned with \textsc{equiv} regularization are again easily the most equivariant.}  We also see that for all methods error is lower for atomic motions than composite motions, since they are more equivariant for smaller motions (see Supp).

\input{pair_nn_fig}

\subsection{Recognition results}\label{sec:recog_results}

Next we test the unsupervised-to-supervised transfer pipeline of Sec~\ref{sec:recog} on 3 recognition tasks: NORB-NORB, KITTI-KITTI, and KITTI-SUN.  The first dataset in each pairing is unsupervised, and the second is supervised.

Table~\ref{tab:recog} (center) shows the results.  On all 3 datasets, our method significantly improves classification accuracy, not just over the no-prior \textsc{clsnet} baseline, but also over the closest previous unsupervised feature learning methods.\footnote{To verify the \textsc{clsnet} baseline is legitimate, we also ran a Tiny Image nearest neighbor baseline on SUN as in~\cite{sun}.  It obtains 0.61\% accuracy (worse than \textsc{clsnet}, which obtains 0.70\%).}

All the unsupervised feature learning methods yield large gains over \textsc{clsnet} on all three tasks.  However, \textsc{drlim} and \textsc{temporal} are significantly weaker than the proposed method. Those methods are based on the ``slow feature analysis'' principle~\cite{Wiskott2002}---nearby frames must be close to one another in the learned feature space. We observe in practice (see Supp)
that temporally close frames are mapped close to each other after only a few training epochs. This points to a possible weakness in these methods---even with parameters (temporal neighborhood size, regularization $\lambda$) cross-validated for recognition, the slowness prior is too weak to regularize feature learning effectively,
since strengthening it causes loss of discriminative information. %

In contrast, our method requires \emph{systematic} feature space responses to ego-motions, and offers a stronger prior.
{\textsc{equiv+drlim} further improves over \textsc{equiv}, possibly because: (1) our \textsc{equiv} implementation only exploits frame pairs arising from specific motion patterns as positives, while \textsc{drlim} more broadly exploits all neighbor pairs, and (2) \textsc{drlim} and \textsc{equiv} losses are compatible--- \textsc{drlim} requires that small perturbations affect features in small ways, and \textsc{equiv} requires that they affect them systematically.
}

The most exciting result is KITTI-SUN.  The KITTI data itself is vastly more challenging than NORB due to its noisy ego-poses from inertial sensors, dynamic scenes with moving traffic, depth variations, occlusions, and objects that enter and exit the scene.  Furthermore, the fact we can transfer \textsc{equiv} features learned without class labels on KITTI (street scenes from Karlsruhe, road-facing camera with fixed pitch and field of view) to be useful for a supervised task
on the very different domain of SUN (``in the wild'' web images from 397 categories mostly unrelated to streets) indicates the generality of our approach.
Our best recognition accuracy of 1.58\% on SUN is achieved with only 6 labeled examples per class.  It is $\approx$30\% better than the nearest competing baseline \textsc{temporal} and over 6 times better than chance. Top-10 accuracy trends are similar.

\newchanges{While we have thus far kept supervised training sets small to simulate categorization problems in the ``long tail'' where training samples are scarce and priors are most useful, new preliminary tests with larger labeled training sets on SUN show that our advantage is preserved. With $N$=20 samples for each of 397 classes on KITTI-SUN, \textsc{equiv} scored 3.66+/-0.08\% accuracy vs. 1.66+/-0.18 for \textsc{clsnet}.}

\vspace{-0.03in}
\subsection{Next-best view selection for recognition}\label{sec:nbv}
Next, we show preliminary results of a direct application of equivariant features to ``next-best view selection''. Given one view of a NORB object, the task is to tell a hypothetical robot how to move next to help recognize the object, \ie, which neighboring view would best reduce object prediction uncertainty. We exploit the fact that equivariant features behave predictably under ego-motions to identify the optimal next view. Our method for this task, similar in \changes{spirit} to~\cite{wu20143d}, is described in detail in Supp. Table~\ref{tab:nbv} (right) shows the results. On this task too, \textsc{equiv} features easily outperform the baselines.

\vspace{-0.02in}
\subsection{Qualitative analysis}\label{sec:qual_analysis}

To qualitatively evaluate the impact of equivariant feature learning, we pose a nearest neighbor task in the \emph{feature difference} space to retrieve image pairs related by similar ego-motion to a query image pair (details in Supp).
Fig~\ref{fig:pairdiff_nn} shows examples.  For a variety of query pairs, we show the top neighbor pairs in the \textsc{equiv} space, as well as in pixel-difference space for comparison.  Overall they visually confirm the desired equivariance property: neighbor-pairs in \textsc{equiv}'s difference space exhibit a similar transformation (turning, zooming, \etc), whereas those in the original image space often do not. Consider the first azimuthal rotation NORB query in row 2, where pixel distance, perhaps dominated by the lighting, identifies a wrong ego-motion match, whereas our approach finds a correct match, despite the changed object identity, starting azimuth, lighting \etc. The red boxes show failure cases. For instance, in the KITTI failure case shown (row 1, column 3), large foreground motion of a truck in the query image causes our method to wrongly miss the rotational motion.

\changes{\section{Conclusion}}
Over the last decade, visual recognition methods have focused almost exclusively on learning from \newchanges{``bags of images".  We argue that} such ``disembodied" image collections, though clearly valuable when collected at scale, deprive feature learning methods from the informative physical context of the original visual experience.  We presented the first ``embodied'' approach to feature learning that generates features equivariant to ego-motion. Our results on multiple datasets and on multiple tasks show that our approach successfully learns equivariant features, which are beneficial for many downstream tasks and hold great promise for novel future applications.

%% file: pair_nn_fig.tex
\begin{figure*}[ht]
  \centering
  \includegraphics[width=1\linewidth]{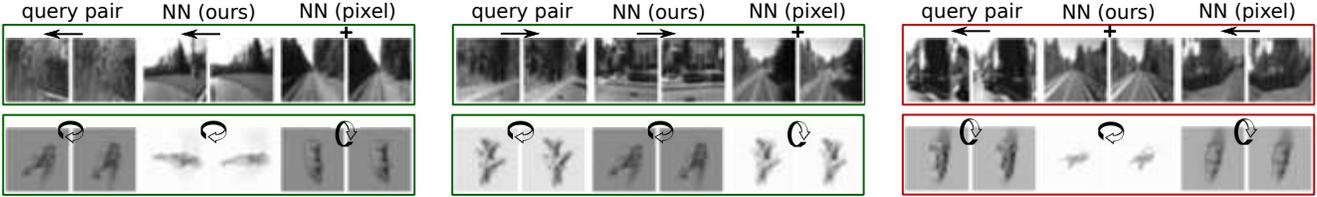}
  \caption{Nearest neighbor image pairs (cols 3 and 4 in each block) in pairwise equivariant feature difference space for various query image pairs (cols 1 and 2 per block). For comparison, cols 5 and 6 show pixel-wise difference-based neighbor pairs. The direction of ego-motion in query and neighbor pairs (inferred from ego-pose vector differences) is indicated above each block.
 See text.}
  \label{fig:pairdiff_nn}
\end{figure*}

%% file: ack.tex
\vspace{0.05in}
\noindent \small{\textbf{Acknowledgements:} This research is supported in part by ONR PECASE Award N00014-15-1-2291 and a gift from Intel.}

%% file: arxiv_supp.tex
\section{Supplementary details}
\subsection{KITTI and SUN dataset samples}
\begin{figure*}[t]
 \centering
  \includegraphics[width=0.8\linewidth]{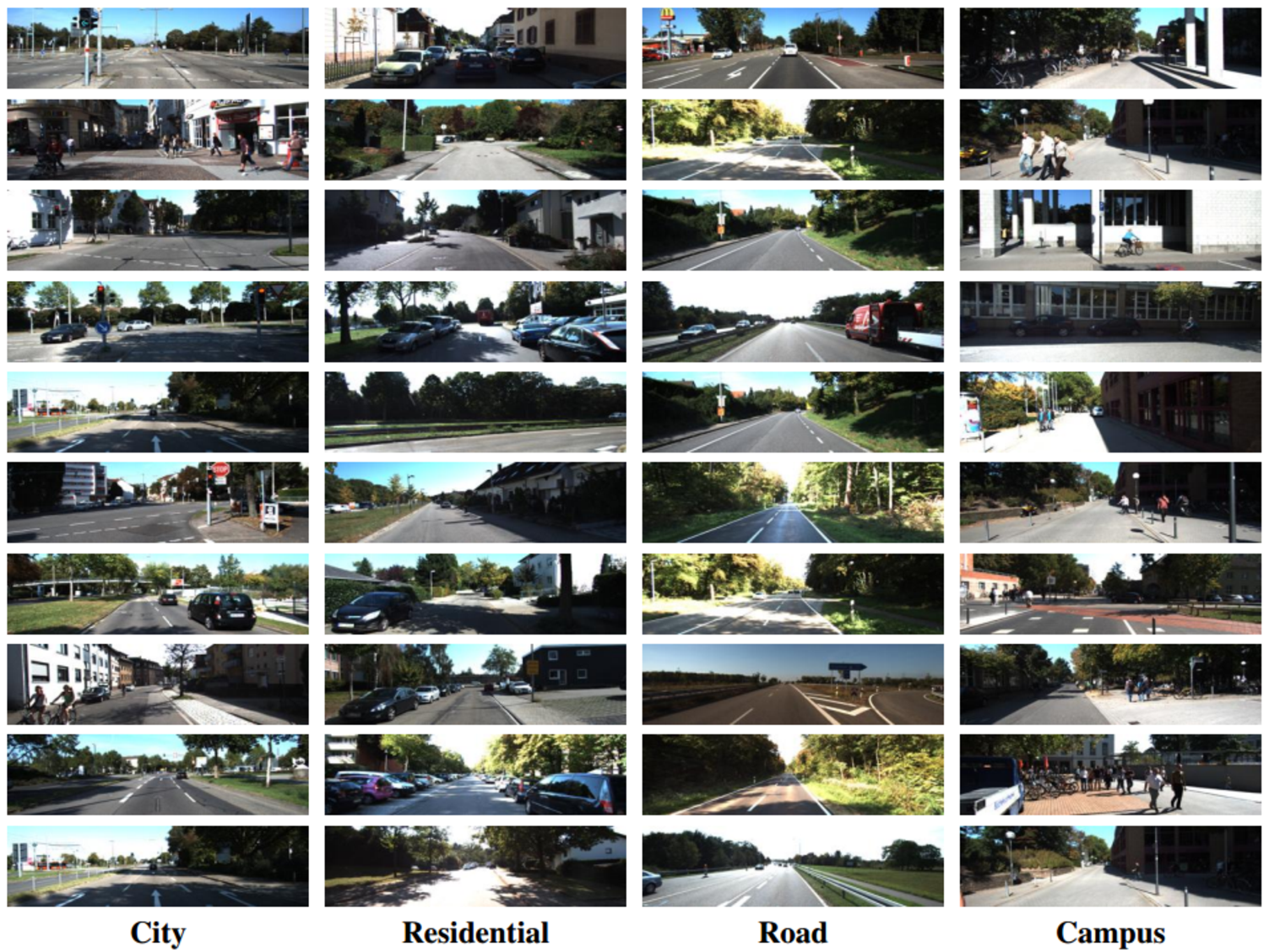}
  \\
  \includegraphics[width=0.8\linewidth]{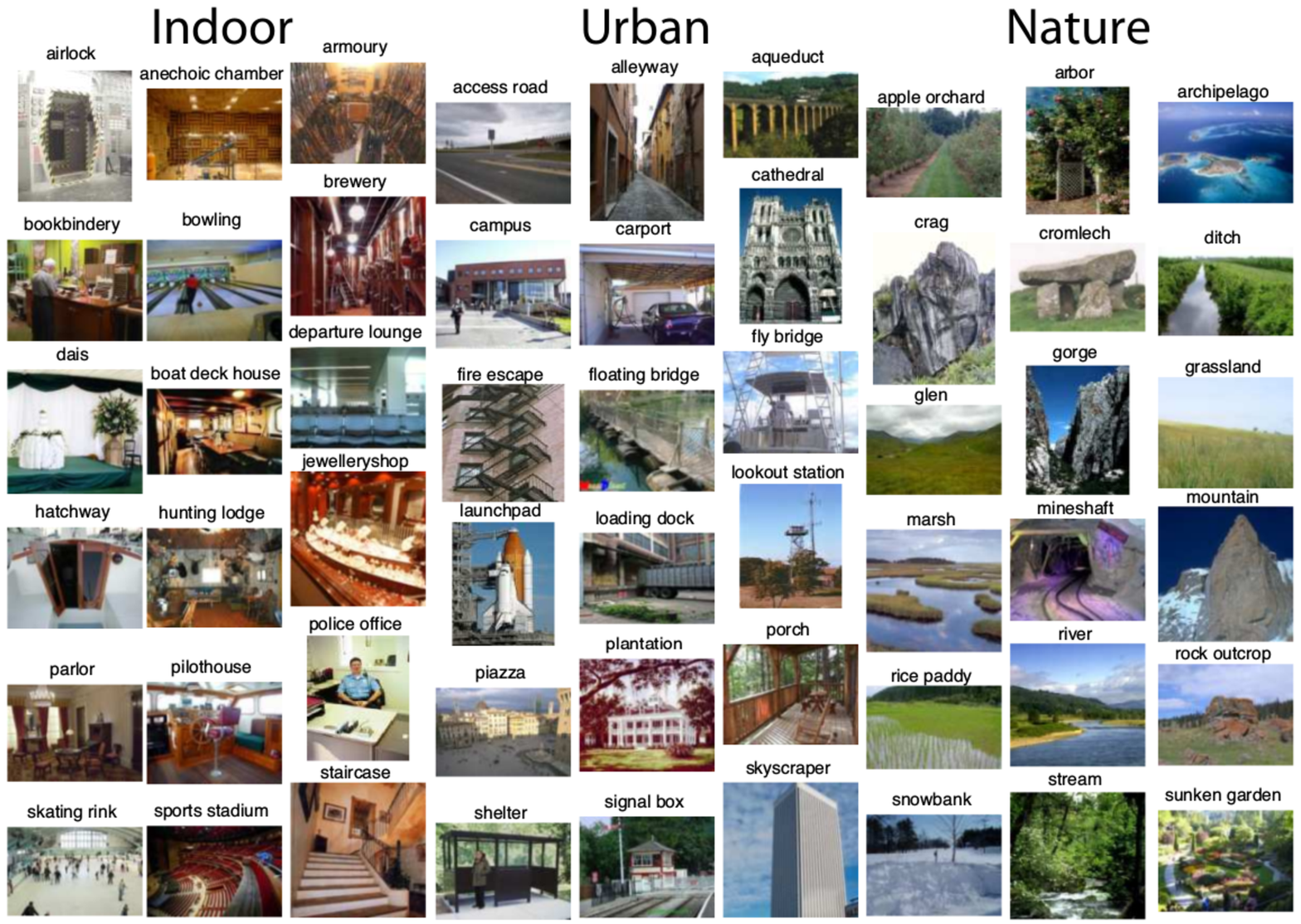}
  \caption{(top) Figure from~\cite{kitti} showcasing images from the 4 KITTI location classes (shown here in color; we use grayscale images), and (bottom) Figure from~\cite{sun} showcasing images from a subset of the 397 SUN classes (shown here in color; see text in main paper for image pre-processing details).}
  \label{fig:kitti_and_sun}
\end{figure*}

Some sample images from KITTI and SUN are shown in Fig~\ref{fig:kitti_and_sun}. As they show, these datasets have substantial domain differences. In KITTI, the camera faces the road and has a fixed field of view and camera pitch, and the content is entirely street scenes around Karlsruhe.  In SUN, the images are downloaded from the internet, and belong to 397 diverse indoor and outdoor scene categories---most of which have nothing to do with roads.

\subsection{Optimization and hyperparameter selection (Main Sec \ref{sec:setup})}
\begin{figure}
  \centering
  \includegraphics[width=1\linewidth]{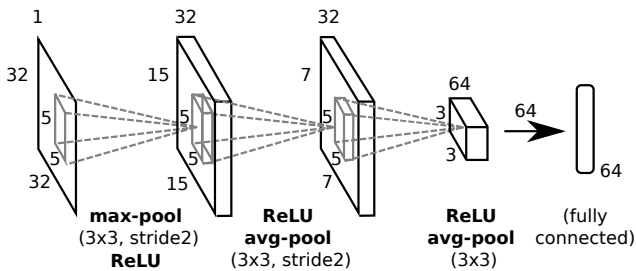}
  \caption{{KITTI $\mathbf{z_{\bm{\theta}}}$ architecture producing $D=$64-dim. features: 3 convolution layers and a fully connected feature layer (non-linear operations specified along the bottom).}}
  \vspace{-0.1in}
  \label{fig:kitti_net}
\end{figure}

(Elaborating on para titled ``Network architectures and Optimization'' \ref{sec:setup})
As mentioned in the paper, for KITTI, we closely follow the cuda-convnet~\cite{cuda-convnet} recommended CIFAR-10 architecture: 32 conv(5x5)-max(3x3)-ReLU $\rightarrow$ 32 conv(5x5)-ReLU-avg(3x3) $\rightarrow$ 64 conv(5x5)-ReLU-avg(3x3) $\rightarrow$ $D=$64 full feature units. A schematic representation for this architecture is shown in Fig~\ref{fig:kitti_net}.

We use Nesterov-accelerated stochastic gradient descent as implemented in Caffe~\cite{caffe}, starting from weights randomly initialized according to~\cite{xavier}. The base learning rate and regularization $\lambda$s are selected with greedy cross-validation.
Specifically, for each task, the optimal base learning rate (from 0.1, 0.01, 0.001, 0.0001) was identified for \textsc{clsnet}. Next, with this base learning rate fixed, the optimal regularizer weight  (for \textsc{drlim}, \textsc{temporal} and \textsc{equiv}) was selected from a logarithmic grid (steps of $10^{0.5}$). For \textsc{equiv+drlim}, the \textsc{drlim} loss regularizer weight fixed for \textsc{drlim} was retained, and only the \textsc{equiv} loss weight was cross-validated.
The contrastive loss margin parameter $\delta$ in Eq~\eqref{eq:contrastive_loss}
 in \textsc{drlim}, \textsc{temporal} and \textsc{equiv} were set uniformly to 1.0.
Since no other part of these objectives (including the softmax classification loss) depends on the scale of features,\footnote{Technically, the \textsc{equiv} objective in Eq~\eqref{eq:loss1} may benefit from setting different margins corresponding to the different ego-motion patterns, but we overlook this in favor of scalability and fewer hyperparameters.} different choices of margins $\delta$ in these methods lead to objective functions with equivalent optima - the features are only scaled by a factor.
For \textsc{equiv+drlim}, we set the \textsc{drlim} and \textsc{equiv} margins respectively to 1.0 and 0.1
to reflect the fact that the equivariance maps $M_g$ of Eq~\eqref{eq:loss1} applied to the representation $\mathbf{z_{\bm{\theta}}}(g\bm{x})$ of the transformed image must bring it closer to the original image representation $\mathbf{z_{\bm{\theta}}}(\bm{x})$ than it was before \ie $\|M_g \mathbf{z_{\bm{\theta}}}(g\bm{x})- \mathbf{z_{\bm{\theta}}}(\bm{x})\|_2 < \|\mathbf{z_{\bm{\theta}}}(g\bm{x}) - \mathbf{z_{\bm{\theta}}}(\bm{x})\|_2$. %

In addition, to allow fast and thorough experimentation, we set the number of training epochs for each method on each dataset based on a number of initial runs to assess the scale of time usually taken before the classification softmax loss on validation data began to rise \ie overfitting began. All future runs for that method on that data were run to roughly match (to the nearest 5000) the number of epochs identified above. For most cases, this number was of the order of $50000$. Batch sizes (for both the classification stack and the Siamese networks) were set to 16 (found to have no major difference from 4 or 64) for NORB-NORB and KITTI-KITTI, and to 128 (selected from 4, 16, 64, 128) for KITTI-SUN, where we found it necessary to increase batch size so that meaningful classification loss gradients were computed in each SGD iteration, and training loss began to fall, despite the large number (397) of classes.

On a single Tesla K-40 GPU machine, NORB-NORB training tasks took $\approx$15 minutes, KITTI-KITTI tasks took $\approx$30 minutes, and KITTI-SUN tasks took $\approx$2 hours.

\subsection{Equivariance measurement (Main Sec \ref{sec:sanitycheck})}\label{sec:equivar}

\paragraph{Computing $\rho_g$ - details}
In Sec~\ref{sec:sanitycheck} in the main paper, we proposed the following measure for equivariance.
For each ego-motion $g$, we measure equivariance separately through the normalized error $\rho_g$:\vspace{-0.05in}
\begin{equation}
  \rho_g=E\left[\frac{\|\mathbf{z_{\bm{\theta}}}(\bm{x})-M_g^\prime \mathbf{z_{\bm{\theta}}}(g\bm{x})\|_2}{\|\mathbf{z_{\bm{\theta}}}(\bm{x})-\mathbf{z_{\bm{\theta}}}(g\bm{x})\|_2}\right],
  \label{eq:equiv_measure}
\end{equation}
where $E[.]$ denotes the empirical mean, $M_g^\prime$ is the equivariance map, and $\rho_g=0$ would signify perfect equivariance. \changes{We closely follow the equivariance evaluation approach of~\cite{Vedaldi2014} to solve for the equivariance maps of features produced by each compared method on held-out validation data (cf. Sec~\ref{sec:setup} from the paper), before computing $\rho_g$.}
Such maps are produced explicitly by our method, but not the baselines.  Thus, as in~\cite{Vedaldi2014}, we compute their maps\footnote{For uniformity, we do the same recovery of $M_g^\prime$ for our method; our results are similar either way.}  by solving a least squares minimization problem based on the definition of equivariance in Eq~\eqref{eq:equivar} in the paper:
\begin{equation}
  M_g^\prime= \argmin_{M} \sum_{m(\bm{y}_i,\bm{y}_j)=g} \|\mathbf{z}_{\bm{\theta}}(\bm{x}_i)-M \mathbf{z}_{\bm{\theta}}(\bm{x}_j)\|_2.
  \label{eq:computing_eq_map}
\end{equation}
$M_g^\prime$'s computed as above are used to compute $\rho_g$'s as in Eq~\eqref{eq:equiv_measure}. $M_g^\prime$ and $\rho_g$ are computed on disjoint subsets of the validation data. Since the output features are relatively low in dimension ($D=$ 100), we find regularization for Eq~\eqref{eq:computing_eq_map} unnecessary.

\paragraph{Equivariance results - details}
While results in the main paper (Table~\ref{tab:equiv_measurement}) were reported as averages over atomic and composite motions, we present here the results for individual motions in Table~\ref{tab:equiv_measurement}. While relative trends among the methods remain the same as for the averages reported in the main paper, the new numbers help demonstrate that $\rho_g$ for composite motions is no bigger than for atomic motions, as we would expect from the argument presented in Sec~\ref{sec:equivar} in the main paper.

To see this, observe that even among the atomic motions, $\rho_g$ for all methods is lower on the small ``up'' atomic ego-motion (5\textdegree) than it is for the larger ``right'' ego-motion (20\textdegree). Further, the errors for ``right'' are close to those for the composite motions (``up+right'', ``up+left'' and ``down+right''), establishing that while equivariance is diminished for larger motions, it is not affected by whether the motions were used in training or not. In other words, if trained for equivariance to a suitable discrete set of atomic ego-motions (cf. Sec~\ref{sec:equivar} in the paper), the feature space generalizes well to new ego-motions.

\begin{table}[t]
  \centering
    \small{
      \scalebox{0.88}{
    \begin{tabular}{|l||c|c||c|c|c|}
    \hline
    Tasks $\rightarrow$                  & \multicolumn{2}{c||}{atomic}       &  \multicolumn{3}{c|}{composite}\\ \hline
    Datasets $\downarrow$                & ``up (u)'' & ``right (r)'' & ``u+r''  & ``u+l'' & ``d+r''\\\hline
    random                               & 1.0000    & 1.0000       &   1.0000 & 1.0000  &  1.0000\\
    \textsc{clsnet}                      & 0.9276    & 0.9202       & 0.9222   & 0.9138  &  0.9074\\
    \textsc{temporal}~\cite{Mobahi2009}  & 0.7140    & 0.8033       & 0.8089   & 0.8061  &  0.8207\\
    \textsc{drlim}~\cite{drlim}          & 0.5770    & 0.7038       & 0.7281   & 0.7182  &  0.7325\\
    \textsc{equiv}                       & 0.5328    & 0.6836       & 0.6913   & 0.6914  &  0.7120\\
    \textsc{equiv+drlim}                 & \textbf{0.5293} & \textbf{0.6335} &\textbf{0.6450} & \textbf{0.6460} & \textbf{0.6565}\\ \hline
    \end{tabular}
}
\caption{The ``normalized error'' equivariance measure $\rho_g$ for individual ego-motions (Eq~\eqref{eq:equiv_measure}) on NORB, organized as ``atomic'' (motions in the \textsc{equiv} training set) and ``composite'' (novel) ego-motions.}
\label{tab:equiv_measurement}
}
\end{table}

\subsection{Recognition results (Main Sec~\ref{sec:recog_results})}

\paragraph{Restricted slowness is a weak prior} We now present evidence supporting our claim in the paper that the principle of slowness, which penalizes feature variation within small temporal windows, provides a prior that is rather weak. In every stochastic gradient descent (SGD) training iteration for the \textsc{drlim} and \textsc{temporal} networks, we also computed a ``slowness'' measure that is independent of feature scaling (unlike the \textsc{drlim} and \textsc{temporal} losses of Eq~\ref{eq:temp_coherence} themselves), to better understand the shortcomings of these methods.

Given training pairs $(\bm{x}_i,\bm{x}_j)$ annotated as neighbors or non-neighbors by $n_{ij}=\mathbbm{1}(|t_i-t_j|\leq T)$ (cf. Eq~\eqref{eq:temp_coherence} in the paper), we computed pairwise distances $\Delta_{ij}=d(\mathbf{z}_{\bm{\theta}(s)}(\bm{x}_i), \mathbf{z}_{\bm{\theta}(s)}(\bm{x}_j))$, where $\bm{\theta}(s)$ is the parameter vector at SGD training iteration $s$, and $d(.,.)$ is set to the $\ell_2$ distance for \textsc{drlim} and to the $\ell_1$ distance for \textsc{temporal} (cf. Sec~\ref{sec:exp}).

We then measured how well these pairwise distances $\Delta_{ij}$ predict the temporal neighborhood annotation $n_{ij}$, by measuring the Area Under Receiver Operating Characteristic (AUROC) when varying a threshold on $\Delta_{ij}$.

These ``slowness AUROC''s are plotted as a function of training iteration number in Fig~\ref{fig:slowness_AUROC}, for \textsc{drlim} and \textsc{coherence} networks trained on the KITTI-SUN task. Compared to the standard random AUROC value of 0.5, these slowness AUROCs tend to be near 0.9 already even before optimization begins, and reach peak AUROCs very close to 1.0 on both training and testing data within about 4000 iterations (batch size 128). This points to a possible weakness in these methods---even with parameters (temporal neighborhood size, regularization $\lambda$) cross-validated for recognition, the slowness prior is too weak to regularize feature learning effectively, since strengthening it causes loss of discriminative information. In contrast, our method requires \emph{systematic} feature space responses to ego-motions, and offers a stronger prior.

\begin{figure}[t]
  \centering
  \includegraphics[width=1\linewidth]{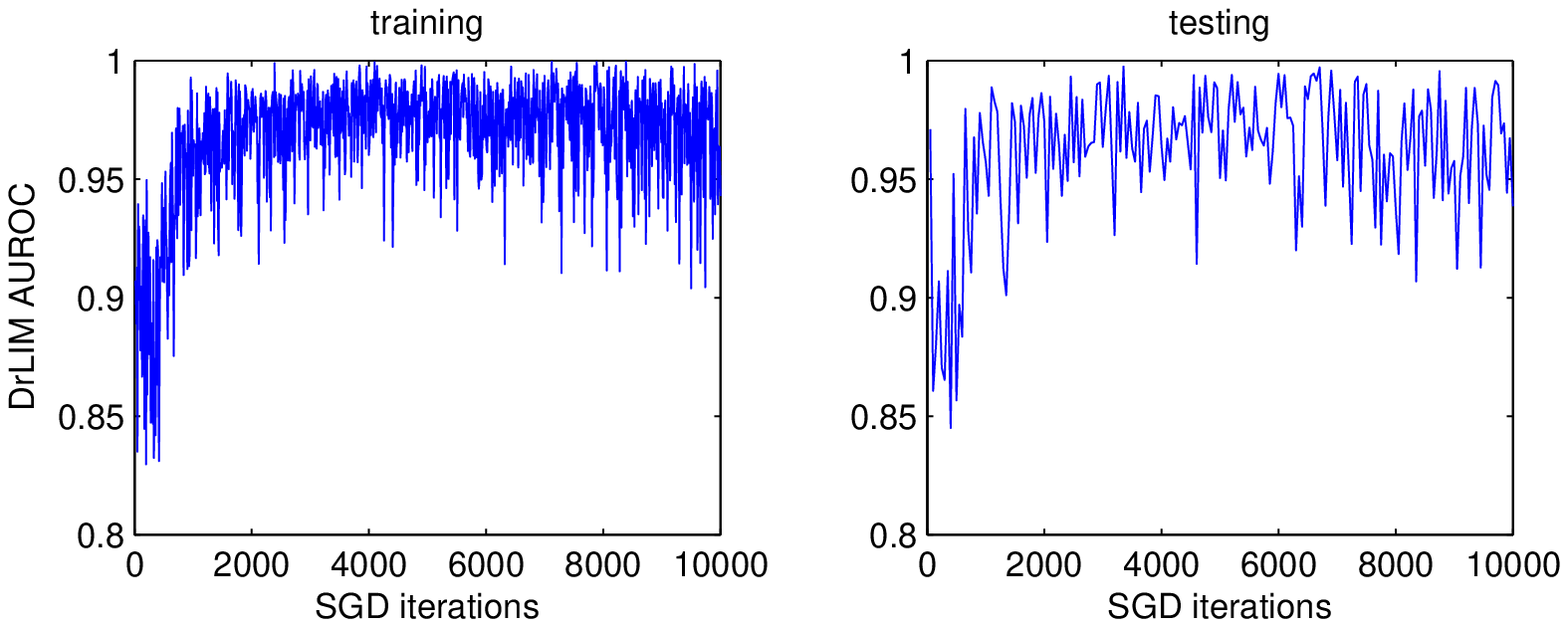}
  \includegraphics[width=1\linewidth]{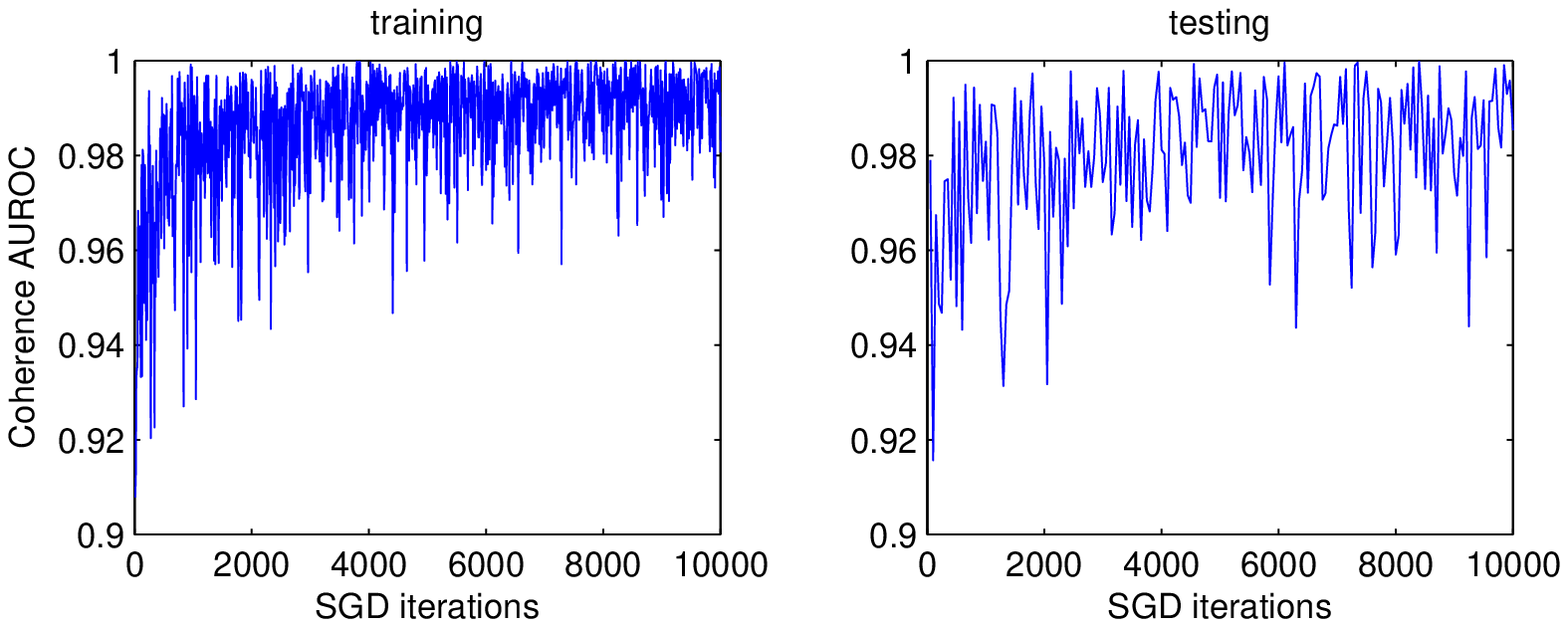}
  \caption{Slowness AUROC on training (left) and testing (right) data for (top) \textsc{drlim} (bottom) \textsc{coherence}, showing the weakness of slowness prior.}
  \label{fig:slowness_AUROC}
\end{figure}
\begin{figure*}[ht]
  \centering
  \includegraphics[width=1\linewidth]{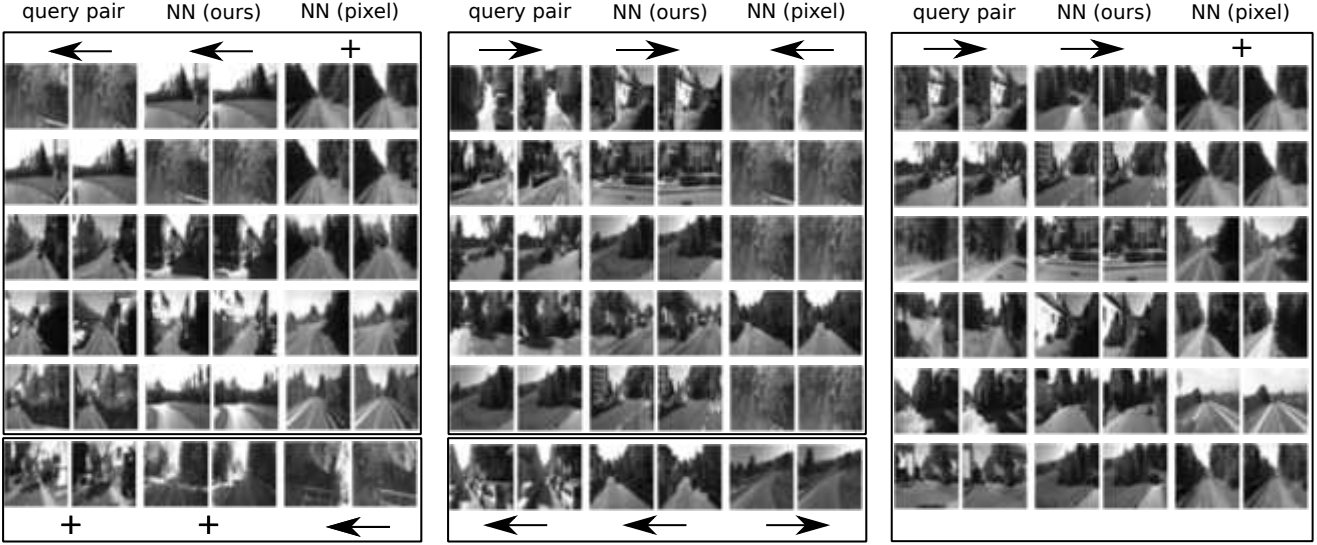}
  \caption{ (Contd. from Fig~\ref{fig:pairdiff_nn}) More examples of nearest neighbor image pairs (cols 3 and 4 in each block) in pairwise equivariant feature difference space for various query image pairs (cols 1 and 2 per block). For comparison, cols 5 and 6 show pixel-wise difference-based neighbor pairs. The direction of ego-motion in query and neighbor pairs (inferred from ego-pose vector differences) is indicated above each block.}
  \label{fig:pair_nn_cases}
\end{figure*}

\subsection{Next-best view selection (Main Sec~\ref{sec:nbv})}
We now describe our method for next-best view selection for recognition on NORB. Given one view of a NORB object, the task is to tell a hypothetical robot how to move next to help recognize the object \ie which neighboring view would best reduce object prediction uncertainty. We exploit the fact that equivariant features behave predictably under ego-motions to identify the optimal next view.

We limit the choice of next view $g$ to $\{$ ``up'', ``down'', ``up+right'' and ``up+left'' $\}$ for simplicity in this preliminary test. We build a $k$-nearest neighbor (k-NN) image-pair classifier for each possible $g$, using only training image pairs $(\bm{x},g\bm{x})$ related by the ego-motion $g$. This classifier $C_g$ takes as input a vector of length $2D$, formed by appending the features of the image pair (each image's representation is of length $D$) and produces the output probability of each class. So, $C_g([\mathbf{z}_{\bm{\theta}}(\bm{x}),~  \mathbf{z}_{\bm{\theta}}(g\bm{x})])$ returns class likelihood probabilities for all 25 NORB classes. Output class probabilities for the k-NN classifier are computed from the histogram of class votes from the $k$ nearest neighbors. We set $k=25$.

At test time, we first compute features $\mathbf{z}_{\bm{\theta}}(\bm{x}_0)$ on the given starting image $\bm{x}_0$. Next we predict the feature $\mathbf{z}_{\bm{\theta}}(g \bm{x}_0)$ corresponding to each possible surrounding view $g$, as $M_g^\prime \mathbf{z}_{\bm{\theta}}(\bm{x}_0)$, per the definition of equivariance (cf. Eq~\ref{eq:equivar} in the paper).\footnote{Equivariance maps $M_g^\prime$ for all methods are computed as described in Sec~\ref{sec:equivar} in this document (and Sec~\ref{sec:sanitycheck} in the main paper)} %

With these predicted transformed image features and the pair-wise nearest neighbor class probabilities $C_g(.)$, we may now pick the next-best view as:
\begin{equation}
g^*=\argmin_g H(C_g([\mathbf{z}_{\bm{\theta}}(\bm{x}_0),~  M_g^\prime \mathbf{z}_{\bm{\theta}}(\bm{x_0})])),
  \label{eq:nbv}
\end{equation}
where $H(.)$ is the information-theoretical entropy function. This selects the view that would produce the least predicted image pair class prediction uncertainty.

\subsection{Qualitative analysis (Main Sec~\ref{sec:qual_analysis})}

To qualitatively evaluate the impact of equivariant feature learning, we pose a pair-wise nearest neighbor task in the \emph{feature difference} space to retrieve image pairs related by similar ego-motion to a query image pair (details in Supp).
Given a learned feature space $\mathbf{z(.)}$ and a query image pair $(\bm{x}_i,\bm{x}_j)$, we form the pairwise feature difference  $\bm{d}_{ij}=\mathbf{z}(\bm{x}_i)-\mathbf{z}(\bm{x}_j)$.
In an equivariant feature space, other image pairs $(\bm{x}_k,\bm{x}_l)$ with similar feature difference vectors $\bm{d}_{kl}\approx \bm{d}_{ij}$ would be likely to be related by similar ego-motion to the query pair.\footnote{\cc{Note that in our model of equivariance, this isn't strictly true, since the pair-wise difference vector $M_g\mathbf{z_{\bm{\theta}}}(\bm{x})-\mathbf{z_{\bm{\theta}}}(\bm{x})$ need not actually be fixed for a given transformation $g$, $\forall\bm{x}$. For small motions (and the right kinds of equivariant maps $M_g$), this still holds approximately, as we find in practice.}}  This can also be viewed as an analogy completion task, $\bm{x}_i : \bm{x}_j = \bm{x}_k : ?$, where the right answer should apply $p_{ij}$ to $\bm{x}_k$ to obtain $\bm{x}_l$. For the results in the paper, the closest pair to the query in the learned equivariant feature space is compared to that in the pixel space. Some more examples are shown in Fig~\ref{fig:pair_nn_cases}.